\documentclass{article}

\PassOptionsToPackage{numbers, compress}{natbib}
\usepackage[final]{neurips_2024}

\usepackage[utf8]{inputenc} % allow utf-8 input
\usepackage[T1]{fontenc}    % use 8-bit T1 fonts
\usepackage{hyperref}       % hyperlinks
\usepackage{url}            % simple URL typesetting
\usepackage{booktabs}       % professional-quality tables
\usepackage{amsfonts}       % blackboard math symbols
\usepackage{graphicx}
\usepackage{multirow}
\usepackage{amsmath}
\usepackage{mathrsfs}
\usepackage{nicefrac}       % compact symbols for 1/2, etc.
\usepackage{microtype}      % microtypography
\usepackage{xcolor}         % colors
\usepackage{authblk}
\usepackage{etoolbox}
\newcommand{\etal}{\textit{et al.}}

\title{ETO:Efficient Transformer-based Local Feature Matching by Organizing Multiple Homography Hypotheses}

\author{%
  \textbf{Junjie Ni}$^{1}$  \quad
  \textbf{Guofeng Zhang}$^{1*}$ \quad
  \textbf{Guanglin Li}$^{1}$ \quad
  \textbf{Yijin Li}$^{1}$ \\
  \textbf{Xinyang Liu}$^{1}$ \quad
  \textbf{Zhaoyang Huang}$^{2}$ \quad
  \textbf{Hujun Bao}$^{1}$\thanks{Corresponding author}  \\ \vspace{0.5cm}
  
  $^{1}$State Key Lab of CAD\&CG, Zhejiang University \quad $^{2}$CUHK MMLab \\
}
\begin{document}

\maketitle

\begin{abstract}
% Local feature matching is a cornerstone task in computer vision. 
% 第一句话和后面的话没有逻辑关系？，感觉很突兀，考虑一下这句话放哪里比较好
Recent developments have led to the emergence of transformer-based approaches for local feature matching, resulting in enhanced accuracy of matches. 
% We observe that the CNN-based network architectures often excel in matching speed, while the transformer-based architectures tend to provide more accurate matches but are too slow.
% This disparity poses challenges for practical real-time applications, such as Simultaneously Localization And Mapping(SLAM), Augmented Reality (AR), automatic driving, or Robotics. 
% Our research bridges this performance gap. 
% We present a new transformer-based method that maintains a comparable performance with the earlier ones while achieving the speed of purely CNN-based techniques. 
However, the time required for transformer-based feature enhancement is excessively long, which limits their practical application. In this paper, we propose methods to reduce the computational load of transformers during both the coarse matching and refinement stages. During the coarse matching phase, we organize multiple homography hypotheses to approximate continuous matches. Each hypothesis encompasses several features to be matched, significantly reducing the number of features that require enhancement via transformers. In the refinement stage, we reduce the bidirectional self-attention and cross-attention mechanisms to unidirectional cross-attention, thereby substantially decreasing the cost of computation. Overall, our method demonstrates at least 4 times faster compared to other transformer-based feature matching algorithms.
Comprehensive evaluations on other open datasets such as Megadepth, YFCC100M, ScanNet, and HPatches demonstrate our method's efficacy, highlighting its potential to significantly enhance a wide array of downstream applications.
\end{abstract}

\section{Introduction}
\label{sec:intro}

Local feature matching~\cite{orb, surf} is a fundamental problem in the field of computer vision and plays a significant role in downstream applications, including but not limited to SLAM~\cite{orb-slam,yang2022vox,tof_slam,hu2024cg,hu2024cp,nis_slam}, 3D reconstruction~\cite{mvn_afm,nr_in_a_room}, visual localization~\cite{hloc,vs_net,splatloc}, and object pose estimation~\cite{rnnpose,gcasp}. However, traditional CNN-based methods~\cite{superpoint,d2net} often fail under extreme conditions due to the lack of a global receptive field, thus meeting failure under dramatic changes in scale, illumination, viewpoint, or weakly-textured scenes.

\begin{figure}[htbp]
\vspace{-1.5cm}
\centering
\begin{minipage}[t]{0.48\textwidth}
\centering
\includegraphics[width=5.5cm]{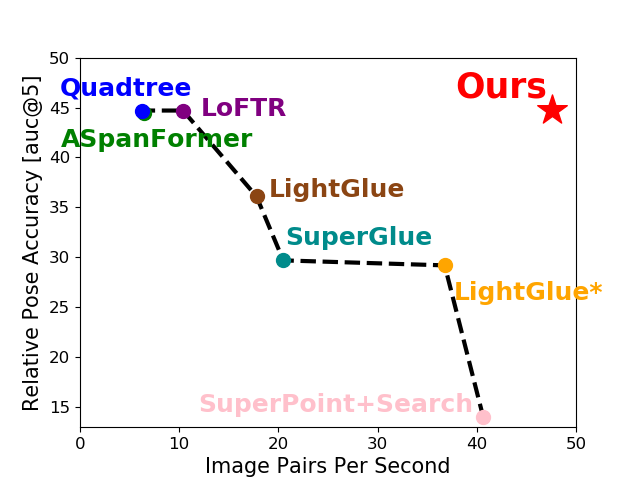}
\caption{\textbf{ETO goes beyond the Pareto curve between accuracy and Efficiency.} This figure shows the performance of different state-of-the-art methods on YFCC100M. We take into account the time for the extraction of feature and their description. LightGlue and LightGlue* are different settings of LightGlue.}
\label{fig:comsheet}
\vspace{-0.5cm}
\end{minipage}
\hspace{0.1cm}
\begin{minipage}[t]{0.48\textwidth}
\centering
\resizebox{1.0\linewidth}{!}{
\includegraphics[width=0.48\textwidth]{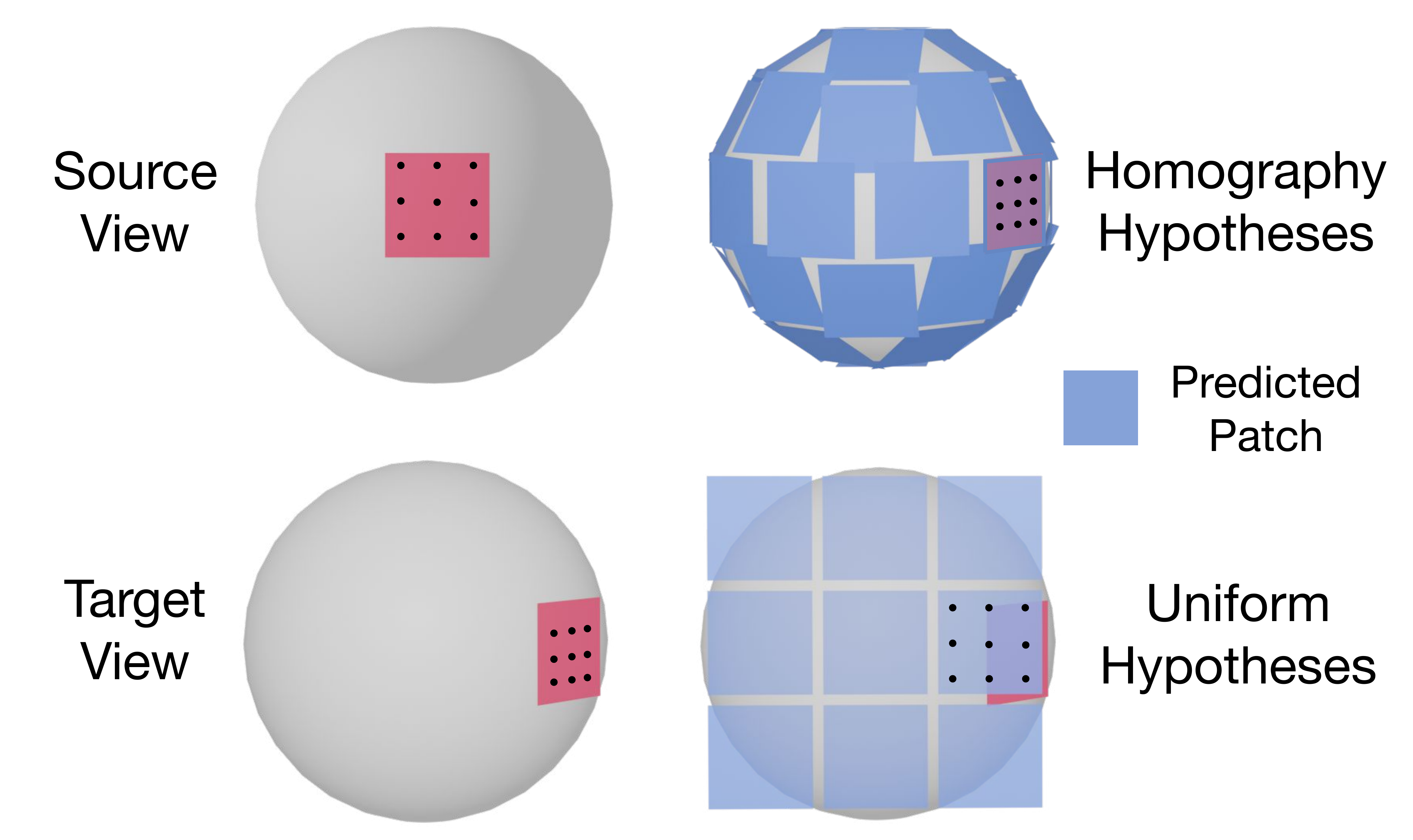}
}
\caption{As demonstrated in the figure, there exists a correspondence between two red regions on the sphere. In contrast to uniform hypotheses, homography hypotheses approximate the correspondence function better, which allows for more precise matching results with fewer computational resources.}
\label{fig:idea}
\vspace{-0.5cm}
\end{minipage}
\end{figure}

% \begin{figure}
% % \vspace{-0.6cm}
% \begin{center}
% \resizebox{1.0\linewidth}{!}{
% \includegraphics[width=0.5\textwidth]{images/sheet.pdf}
% }
% \end{center}
% % \vspace{-0.7cm}
% \caption{\textbf{ETO goes beyond the Pareto curve between accuracy and Efficiency.} This figure shows the performance of different state-of-the-art methods on YFCC100M with RTX2080ti. In this comparison, we take into account time for the extraction of feature points and their description.}
% \label{fig:comsheet}
% \vspace{-0.5cm}
% \end{figure}

% Recently, transformer-based methods~\cite{pats,aspanformer,loftr} forgo traditional CNN-based approaches in favor of integrating global feature maps for matching, significantly alleviating the challenges associated with these cases. Yet, these methods build their foundation highly on the transformer structure~\cite{transformerLindenberger} between a large number of homogenized units, which introduces tremendous computational cost~\cite{superglue,loftr} compared to traditional CNN-based methods~\cite{superpoint},  while the situation is not changed that its matching accuracy is highly affected by number of matching units. Therefore, although past transformer-based researches~\cite{pats,superglue,loftr,quadtree}  have demonstrated significant potential in enhancing the accuracy of local feature matching, the capacity to improve efficiency has not been equally developed.

Recently, some methods~\cite{pats,aspanformer,loftr} forgo traditional CNN-based approaches and base on Transformer~\cite{transformer} for better modeling the long-range dependencies.
However, Transformer is widely known for its high computational complexity especially when it is applied in vision tasks where the computational complexity grows quadratically in the number of the input image tokens (i.e., patches).
To reduce the inherent complexity associated with the Transformer, these methods generally adopt the coarse-to-fine strategy and incorporate more computationally efficient variants of the Transformer, such as the Linear Transformer~\cite{linear_transformer}. Nevertheless, the computational overhead remains substantial and severely hinders the application demanding low-latency operations, such as tracking~\cite{tracking,blinkvision,context_tap}, or those requiring the processing of extensive datasets, such as large-scale mapping~\cite{large_scale}.

In this paper, we propose to solve the efficiency problem of transformer-based local feature matching.
Our insights are twofold. First, we propose to introduce the homography hypothesis in the pipeline. The homography hypothesis is a kind of piece-wise smooth prior to the scene that has long been explored in the vision tasks~\cite{patchmatch}. It allows us to create larger patches and reduce the tokens number that need to be processed in the Transformer. However, it is non-trivial since the regular shape introduced by the homography hypothesis can bring significant errors, especially along the boundary. Besides, how to supervise the training of multiple homography hypotheses with the absence of ground truth remains a problem.
Second, We empirically find it is redundant that the previous methods employ multiple self- and cross-attention in their fine-level stage since the coarse-level stage has conducted sufficient propagation. As a result, the computation complexity can be further reduced.

Specifically, we propose ETO, the Efficient Transformer-based Local Feature Matching by Organizing Multiple Homography Hypotheses. ETO follows previous methods~\cite{loftr, aspanformer} and employs a two-stage coarse-to-fine pipeline.
It first establishes matches at the patch level and then refines the matches to the sub-pixel level.
In the first stage, ETO learns to predict a set of hypotheses, each encompassing multiple patches to be matched. We approximately assume that each patch to be matched within one hypothesis is on the same plane, and thus describe these matches under the homography transformation, as illustrated in Fig.~\ref{fig:idea}. The homography hypotheses allow us to reduce the image tokens (patches) that are fed to the Transformer. For a typical image with a resolution of 640 $\times$ 480, Previous methods feed 80 $\times$ 60 
 tokens to the transformer with 1/8 resolution, while we only need to feed 20 $\times$ 15 with 1/32 resolution, which brings a significant speed up.
To reduce the possible error due to the regular shape of the homography hypotheses, ETO subdivides the patches into multiple sub-patches and re-selects the correct hypothesis for each sub-patch. We model the problem of re-selection as a segmentation problem~\cite{yolov3}.
After that, ETO refines the matches in the second stage.
Unlike previous methods that employ multiple self- and cross-attention, ETO only conducts one cross-attention, and the size of query tokens it use is much smaller than previous methods. We call it uni-directional cross-attention.
Empirically we find uni-directional cross attention converges significantly faster at training while providing much higher efficiency.
As shown in Fig.~\ref{fig:comsheet}, ETO outpaces existing methods, achieving 4-5 times faster than LoFTR~\cite{loftr} and 2-3 times more rapid than LightGlue~\cite{lightglue} while maintaining a comparable accuracy with them.

Our contributions can be summarized as follows.
1) We introduce multiple homography hypotheses for the local feature matching problem, which can greatly compress the number of tokens involved in the transformer.
2) We introduce uni-directional cross-attention in the refinement stage. This structure provides fast inference efficiency while maintaining accuracy.
3) Our method not only matches the performance of other transformer-based approaches on diverse open-source datasets such as Megadepth~\cite{megadepth}, YFCC100M~\cite{yfcc100m}, ScanNet~\cite{scannet}, and HPatches~\cite{hpatches}, but it also operates at a significantly higher speed, outpacing all compared methods.

\section{Related Works}
\label{sec:related}
\textbf{Transformer-based Local Feature Matching.}
To find sparse correspondence between two images under diverse viewpoint movement conditions, traditional hand-crafted~\cite{sift,orb,surf}  or early learning-based approaches methods~\cite{d2net,r2d2} usually match keypoints~\cite{superglue, goodcorr}  with their descriptors after detecting them~\cite{superpoint}.
These kinds of matching pipelines are highly dependent on the description of unique feature points, and they fail naturally at dramatic viewpoint changes or poorly textured scenes.
Thanks to transformer~\cite{transformer}, researchers now have a toolbox to enhance the feature descriptors with global information.
Earlier researches~\cite{superglue} integrate the information that where other key points are to each descriptors of key points.
Then, ~\cite{cotr,ecotr} modeling the mapping relationship as a continuous 2-d function.
Sun~\etal~\cite{loftr} constructs a global matching pipeline for each unit on the feature map with Transformer~\cite{transformer,deltar,flowformer,blinkflow}.
The following works refine this pipeline with optical flow~\cite{aspanformer} or more efficient attention structure~\cite{quadtree}. 
There exists other attempts.~\cite{pdcnet++,dkm,rgm} follow another technical route without transformer. They try to merge the gap between optical flow~\cite{raft,flowformer} and local feature matching with the concept of confidence.
Our approach is based on a similar feature extractor of~\cite{loftr}, while we parameterize more information for the units on the feature map, and finally extend the correspondence relationship to the homography relationship between them. The concept of parameterized units on the feature map for local feature matching is introduced by Ni~\etal~\cite{pats}, but they parameterize the units only with scale.

\noindent\textbf{Paramerterization in Local Feature Matching.}
Conventional techniques, as demonstrated in previous works~\cite{sift,aslfeat,r2d2}, when confronted with appearance differences due to changes in viewpoint, try to construct feature descriptors which are invariant to these changes, involving scale, normal and rotation, etc. For rule-based methods~\cite{sift}, they create hand-crafted descriptors with scale space analysis~\cite{scalespace}. For learning-based methods~\cite{superpoint}, they input many image pairs with different viewpoints, letting the neural network learn about the invariance of these changes in appearance.
Nevertheless, to fully mitigate the impact of these appearance changes, it is essential to accurately estimate their effects. Recent efforts~\cite{cotr,pats,scalenet} have attempted to directly estimate the scale differences between images and resize the images to enhance the refinement stage of feature matching. In a similar route, ~\cite{occ2} has focused on directly estimating rotation to calibrate the local features.
Unlike the methods aimed solely at discovering more invariant feature descriptors, we argue that parameterized local appearance changes not only pose a challenge in finding accurate matches but also provide a direct avenue to achieve more accurate and more efficient matches over a broader spatial extent. It allows us to locally parameterize the image to many planes, thus creating multiple homography relationships, not just point-to-point correspondence.

\noindent\textbf{Acceleration in Local Feature Matching.}
In the realm of local feature matching based on the Transformer architecture, three predominant strategies have been mainly employed to enhance computational efficiency in the past. The first approach involves substituting the Softmax function for the Optimal Transport algorithm introduced by~\cite{superglue}. The second approach seeks to replace the full Transformer with a linear Transformer~\cite{linear_transformer}. LoFTR~\cite{loftr} incorporates both of these strategies, however, it introduces a larger number of matching units compared to SuperGlue~\cite{superglue}, resulting in a considerably slower performance than ~\cite{superglue}.
The third strategy focuses on reducing the number of layers and units within the Transformer. LightGlue~\cite{lightglue} introduces early termination~\cite{universal, adaptive_transformer} and progressive unit selection strategies to accelerate computation, yielding significant improvements in speed. However, the performance of LightGlue heavily relies on SuperPoint~\cite{superpoint}, which puts a ceiling on its acceleration.
In contrast, ETO relies on a more precise parameterized model, achieving higher coarse matching accuracy with a feature map whose resolution is 16 times smaller than LoFTR~\cite{loftr}, which distinguishes ETO from another concurrent work Efficient LoFTR~\cite{eloftr} that accelerates the algorithm with adaptive token selection on the feature map of the same resolution. Furthermore, during the fine matching stage, we introduce a uni-directional cross-attention mechanism, allowing us to achieve higher matching speed while sacrificing only a minimal amount of fine stage accuracy.

\begin{figure*}
\vspace{-2.0cm}
\begin{center}
\includegraphics[width=0.98\textwidth]{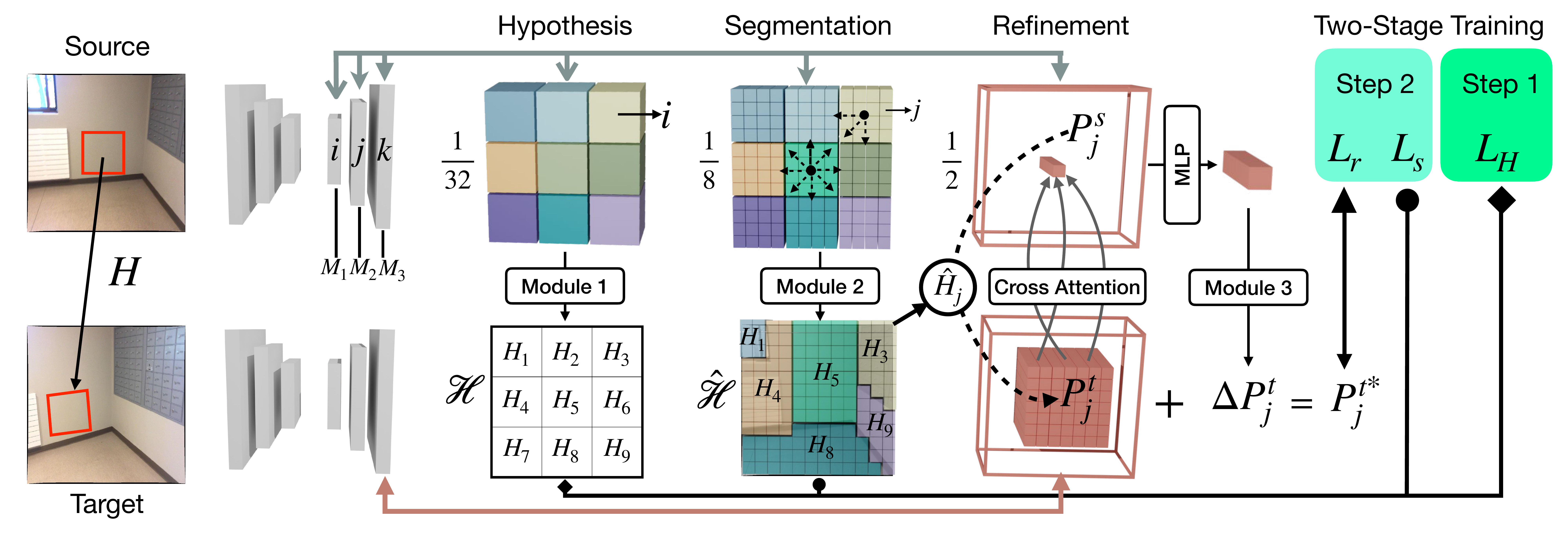}
\end{center}
\vspace{-0.6cm}
   \caption{Given the source image $\mathcal{S}$ and target image $\mathcal{T}$, we first use a U-Net like feature extractor to get images' feature map at different resolution: $M_1$~(H/32~$\times$~W/32), $M_2$~(H/8~$\times$~W/8) and $M_3$~(H/2~$\times$~W/2). We use local $3\times 3$ patches to illustrate our method:~
   \textbf{(a)}~We estimate homography hypotheses $H_i$ for every feature after performing transformer. \textbf{(b)}~We segment the map from these hypotheses to minimize projection errors. With the applied homography matrix $\hat{H}_j$, we can project the chosen source point $P_j^s$ to target point $P_j^t$ . \textbf{(c)}~We update the $P_j^t$ to $P_j^{t*}$ after a uni-directional cross attention. The training process is split into two parts, the coarse and the fine. We train the coarse part with $L_H$, while training the fine part with $L_s$ and $L_r$.}
\vspace{-0.5cm}
\label{fig:pipeline}
\end{figure*}

\section{Method}
\label{sec:Method}

Fig. \ref{fig:pipeline} presents our comprehensive feature matching process, organized into three structured modules. 
These modules are interconnected through feature extractors inspired by U-Net~\cite{unet} and local attributes generated by neural networks. For each unit $i$ on a H/32 $\times$ W/32 resolution feature map $M_1$, we estimate the attributes of homography hypotheses $H_i$. For each unit $j$ on  H/8$\times$W/8 resolution feature map $M_2$, it re-selects the optimal homography hypotheses $\hat{H}_j$ from nearby 9 hypotheses to minimize the projection errors. For the chosen unit $k_j$ on a H/2$\times$W/2 resolution feature map $M_3$, we fix its center point $P^s_j$ at the source image and refine the coordinates of its projected point $P^t_j$ at target image according to $\hat{H}_j$, then get the final matches $P^{t*}_j$. We introduce the feature extractor in Sec. \ref{sec:Feature_extractor}, Sec. \ref{sec:hypotheses} details the estimation of the hypotheses. In Sec. \ref{sec:segmentation} we describe the segmentation of the feature map, Sec. \ref{sec:refinement} delves into refining the matches, while Sec. \ref{sec:supervision} states our supervision methodology.
%-------------------------------------------------------------------------
\subsection{Feature Extraction}
\label{sec:Feature_extractor}
Following~\cite{loftr}, we use ResNet-18~\cite{resnet} as the basic feature extractor to get the feature map with the resolution of $N = \text{H}/32 \times \text{W}/32 $, while we assume that the resolution of the source image and target image is the same. 
Here we get $N = \text{H}/32 \times \text{W}/32 $ \ features and then perform stacked self-attention and cross attention layers between these $N$ tokens to compute the feature map $M_1$. Although there will be $N$~(more than 9) possible patches, our method is mainly performed on locally adjacent patches, so we will omit N in the future and take the local 3$\times$3 patches to illustrate our method (as shown in Fig.~\ref{fig:pipeline}).
Then, we follow~\cite{pats} to upsample the feature map $M_1$ with a U-Net~\cite{unet} like structure.
So, we can obtain the feature maps $M_2$ and $M_3$ at $1/8$ and $1/2$ scale, respectively.

%-------------------------------------------------------------------------
\subsection{Hypothesis Estimation}
\label{sec:hypotheses}
Traditional semi-dense feature matching methods~\cite{loftr,aspanformer} often divide an image into thousands of units. For each unit, they perform bipartite graph matching~\cite{sinkhorn}.
Contrarily, we argue that bipartite graph matching can be extended to the local homography transformation as hypotheses that cover multiple units to be matched. This approach's merit lies in two folds: achieving more precise matches estimated during the first stage and reducing the number of units which are involved in transformer. 

For each unit $i$ on $M_1$ of source images, it is equipped with a feature $f_i^{1}$, a confidence score $c_i$, and a set of hypothesis homography parameters~$H_i$~(including the source positions $p_i^s \in \mathbb{R}^2$, target positions $p_i^t \in \mathbb{R}^2$, rotation $r_i \in \mathbb{R}^1$, scale $s_i \in \mathbb{R}^1$ and  perspective $q_i \in \mathbb{R}^4$). And the unit on target images is indicated by $a$. 
% These attributes for unit $i$ construct a hypothesis, which is denoted as . 

\noindent\textbf{Homography Matrix.}
Initially, we outline the methodology for estimating each unit's local attributes $(p_i^s,p_i^t,r_i,s_i,q_i,c_i)$ and subsequently use these attributes to formulate the local homography matrix. Among these local attributes, the scale $s_i$, rotation $r_i$, and perspective $q_i$ are more related to the feature itself and are regressed directly from features $f_i$ on the source image through an MLP network. 
In contrast, target coordinates $p_i^t$, and confidence scores $c_i$ are more related with the feature map of target images. They are acquired by first identifying the unit  $a_i^*$ with maximal similarity among all target units, and the similarity is defined as the cosine similarity of initial features $f$ on the source image and target image.
%while the similarity matrix $C^i_a$ is defined as cosine similarity between $f_i$ on the source image and on the target image. 
Following this, we construct new features $\hat{f}$ by executing the group-wise correlation~\cite{group_correlation} within the neighborhood of the target units on $M_1$. 
%After that, we use a MLP to extract the centroid coordinates $p_i^t$ and confidence scores $c_i$:
\begin{equation}
\begin{aligned}
  \hat{f}_i &= \mathop{\oplus}\limits_{\delta \in Neighbor(a_i^*)}(<f^1_i, f^1_{\delta}>_g).
  \label{eq:attribute_estimation}
\end{aligned}
\end{equation}
where $<*, *>_g$ is the group-wise correlation~\cite{group_correlation}, and the group size in our method is 8, $Neighbor$ represents the 5$\times$5 neighborhood of unit $a_i^*$,
$\oplus$ indicates the operation of concatenation. 
With the new features $\hat{f}$, 
we use an MLP to process it to get the target position $p_i^t$ and confidence $c_i$.
With these attributes, in order to compute the homography matrix, we establish four target points $B_i^t$ that correspond to four predetermined reference points:
\begin{equation}
\begin{aligned}
  B_i^t &= p_i^t + \mathscr{R}(\mathscr{P}(B_i^s, q_i), r_i) * s_i.
  \label{eq:construct_homography}
\end{aligned}
\end{equation}
Here $B_i^s$ are four imaginary points on source image, while $B_i^t$ are the corresponding target points of $B_i^s$ on the target image. $B_i^t$ is computed from $B_i^s$ by following operation:~$\mathscr{R}$ is the operation of rotation with parameter $r_i$, $\mathscr{P}$ is the perspective transformation operation with parameter $q_i$. These operations allow each variable within the homography matrix $H_i$ to be deduced from 8 projection equations of 4 correspondence $B_i^t = H_i B_i^s$. Details regarding the specific implementation methods for rotation and perspective transformations will be included in the supplementary materials.

%-------------------------------------------------------------------------
\subsection{Segmentation}
\label{sec:segmentation}
\begin{figure}[t]
\vspace{-1.5cm}
\begin{center}
\resizebox{1.0\linewidth}{!}{
\includegraphics[width=0.5\textwidth]{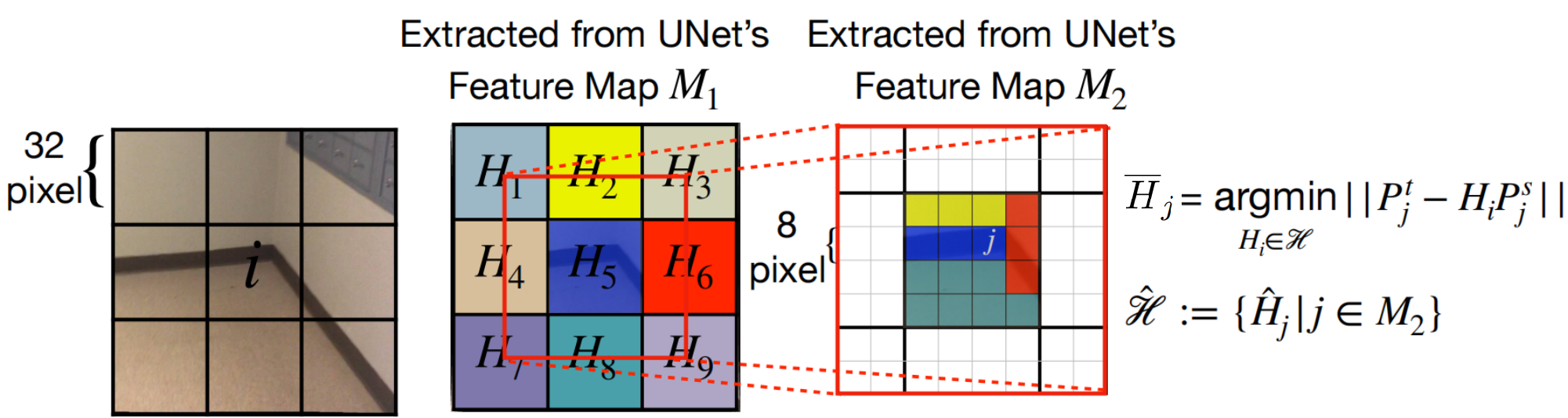}
}
\end{center}
\vspace{-0.5cm}
\caption{Any unit $j$ on $M_2$ should be classified for a hypotheses in $\mathcal{H}$ to minimize projection error. Each $H_i$ describes a plane.}
\label{fig:segmentation}
\vspace{-0.5cm}
\end{figure}

To propagate the homography hypotheses predicted by $M_1$ to a more detailed resolution. We introduce a segmentation operation at the feature map $M_2$ with the resolution $\text{H}/8 \times \text{W}/8$. Segmentation is a per-unit classification task, and we predict a class for each unit $j$ on $M_2$. Here, we only consider locally adjacent $3 \times 3$ patches, and all possible classes is defined as $\mathcal{H}=\{H_i | i=1...9\}$.
This classification~(segmentation) involves that, for each unit $j$, selecting a hypothesis from $\mathcal{H}$ that minimizes the projection error at the center of unit $j$.
%which necessitates a classifying approach to be proposed. 
%Each unit $i$ on $M_1$ encompasses 16 sub-units $j$ on $M_2$, and here we call the hypothesis of $i$ and its 8-neighborhood as $\mathcal{H}_j$. 
% With the previously predicted homography hypotheses, we can only assume the 16 sub-units within $i$ are on the same plane and follow the same hypotheses $H_i$, which may be inaccurate. 
After our proposed segmentation stage, each sub-unit $j$ can find the hypotheses $\hat{H}_j$ that make its error smallest in all possible hypotheses $\mathcal{H}$. We illustrate the intuitive process of segmentation at Fig.~\ref{fig:segmentation}.

Our proposed \textbf{segmentation} differs from traditional semantic segmentation. Instead of aiming for a specific semantic category, it targets a dynamic geometric relationship.
To find the relationship, we introduce a new cosine similarity matrix $C_j$ between the local feature $f^2_j$ on $M_2$ and all features $f^1_i$ on $M_1$, %where $i^n_j$ is the unit corresponds to the alternative hypotheses $H_j^n$. 
Positional encoding is employed during this phase to enhance local features, which is indispensable here because the hypothesis in $\mathcal{H}$ are not equivalent. 
%For example, the hypothesis in the center $H_5$ should inherently carry a higher selection probability.
To predicting the class $\hat{H}_j$ by finding the maximum $C_j$, we generate the computed groundtruth $\overline{H}_j$ as follows:

\begin{equation}
\begin{aligned}
  %H_j &= H_j^{\hat{n}}, \quad \hat{n} = \max\limits_{n} C_j^n, \\
  \overline{H}_j &= \mathop{argmin}\limits_{H_i \in \mathcal{H}} {||P_j^t -H_i P_j^s||}.
\label{eq:segmentation}
\end{aligned}
\end{equation}
where 
%$C_j^n$ is defined by the cosine similarity between $f_j + pos_1(j)$ and $f_{i^n_j} + pos_2(n)$. $pos_1(j)$ and $pos_2(n)$ are different learning based positional encoding functions, $n \in \{0,1,...,8\}$.
%$H_j$ is the hypothesis applied and transform $P_j^s$ to $P_j^t$, while
$\overline{H}_j$ is the optimal hypothesis that minimize the projection error. Then, we use focal loss~\cite{focal_loss} $L_s$ to minimize the segmentation error between predicted $\hat{H}_j$ and the computed groundtruth $\overline{H}_j$. The probability of focal loss is set to $C_j$. 
%and $\gamma$ is set to 2. 

\subsection{Refinement}
\label{sec:refinement}
Following ~\cite{loftr}, to enhance efficiency, only one of the points within each unit $j$ is selected for refinement, which we denote the source and target point as $P_j^s$ and $P_j^t$.
Given $P^t_j = \hat{H}_j P^s_j$, the refinement stage finds the offset $\Delta P^t_j$ of each target point $P^t_j$ relative to a fixed source point $P^s_j$. With the feature $f^3$ from $M_3$, conventional techniques unfold features $f^{3}_k$ from local regions in both source image and target image, followed by self-attention and cross-attention. We claim that this process is unnecessarily slow. Here we eliminate self-attention and reduce cross-attention from a bi-directional process to a uni-directional one. Specifically, the feature $\hat{f}^3_{j}$ is computed by querying the features $f^3_k$ within the neighborhood of the original target point $P_j^t$, and become the final feature vectors $\hat{f}^3_j$ for the point $P_j$ on the images. We illustrate this process in supplementary materials. With the proof of experiments in Sec.~\ref{sec:ablation}, we find that refining a single feature in the local region of $M_2$ is enough to get expected results.
%We find that refine a single target point $P_t^s$ is enough for all points with the same assumptions
% It is meaningless to process features on $M_3$ in the neighborhood of $f^3_{k}$ with the same operation. 
Our findings indicate that this approach can largely diminish the computational load of attention mechanisms while still preserving highly accurate matching outcomes.

% \begin{figure}[t]
% % \vspace{-0.3cm}
% \begin{center}
% \resizebox{0.8\linewidth}{!}{
% \includegraphics[width=0.2\textwidth]{images/lightpats_attention.pdf}
% }
% \end{center}
% \vspace{-0.7cm}
% \caption{Previous methods~\cite{loftr} apply self-attention and cross-attention to each feature within a 5$\times$5 feature map, resulting in 2,500 (25$\times$25$\times$4) inner product calculations to gather feature information within a 4-pixel radius. In contrast, our approach conducts a uni-directional cross-attention solely at the query position on a 7$\times$7 feature map, requiring just 49 inner product calculations to capture feature information up to a 6-pixel distance. This makes our method approximately 50 times faster than the previous approach.}
% \label{fig:attention}
% \vspace{-0.5cm}
% \end{figure}

Finally, following~\cite{yolov3}, we process one fixed element of the final feature vector $\hat{f}^3_{j}$ via cross-attention as the confidence score $c_j$ for the corresponding set of matches. It is supervised by if the error final match is larger than a threshold. The coordinates for the matched pairs on the source and target images are $(P^s_j, P^{t*}_j)$. Therefore we can define the supervision of refinement as:
\begin{equation}
\begin{aligned}
  L_r = |P^{t*}_j - \overline{P}^t_j|_2 + BCE(c_j).
  \label{eq:refinement}
\end{aligned}
\end{equation}
where BCE is binary cross entropy~\cite{bce}, which is a commonly utilized loss function for binary classification problems. Here we use BCE to recognize reliable matches. Here $\overline{P}^t_j$ is the ground truth value of $P^{t*}_j$.

\subsection{Supervision}
\label{sec:supervision} 
\noindent\textbf{Indirect supervision for the homography hypothesis.}
Instead of supervising the attributes of the hypothesis directly, our approach employs indirect supervision by monitoring the correspondences of sampling points that are linked via the homography transformation. 
This design offers the advantage of leveraging an excessive number of ground truth matches to efficiently train a network focused on estimating a set of homography parameters. 
Using the ground truth camera pose and depth from datasets, we can get real matched points $\overline{P}^s$ in source and $\overline{P}^t$ in target images. We sample the matched points to train our method. For each points $p$ in $\overline{P}^s$ , we select only 3$\times$3 adjacent hypotheses $\mathcal{H} = \{H_1, ..., H_9\}$ around it on $M_1$. 
where $H_5$ is in the hypothesis region's center and represents the hypothesis of the region containing point $p$. 
Similar to the necessity of segmentation stated in Sec.~\ref{sec:segmentation}, direct supervision which applies $H_5$ to every $p$ could result in avoidable errors, which arise from the mismatch between the irregular boundaries of planes in the real world and the grid-structured unit $i$ on the source image. 
However, given that these sampling points are merely an auxiliary tool for the loss function, we can directly utilize the ground truth coordinates of the matches to supervise and eliminate classifying each sampling point at this stage.
For each point $p$, we assume that it satisfies a certain homography transformation $\overline{H}_p$, and $\overline{H}_p$ satisfies the following defination:
\begin{equation}
\begin{aligned}
  % e_p &= |(H_p \overline{P}^s_p) - \overline{P}_p^t|_1, \\
  % \hat{n}_p &= \min\limits_{n} e_p^n.
  \overline{H}_p &= \mathop{argmin}\limits_{H_i \in \mathcal{H}} {|H_i p^s - p^t|_1}.
  \label{eq:hpdefine}
\end{aligned}
\end{equation}
%In other words, for any hypothesis $H_i$, its maximum scope is confined to its neighborhood on $M_1$. This selection methodology is illustrated in the ensuing formula:
We denote $p$ in the source image as $p^s$ and target image as $p^t$, we use the following error to optimize our method:
\begin{equation}
\begin{aligned}
  e_p &= |\overline{H}_p p^s - p^t|_1. \\
  % \hat{n}_p &= \min\limits_{n} e_p^n.
  \label{eq:selection_method}
\end{aligned}
\end{equation}
where $|*|_1$ is the L1 norm error. 
%$\hat{P}_p^t$ is the ground truth of target point $P^t_p$. $H_p$ is the chosen hypotheses for $\overline{P}^s$. Here $H_p^5 = H_i$ should be prioritized because $i$ spatially covers the sampling point $p$, so it is considered half of the actual value.

\noindent\textbf{Classification or Correspondence Loss for Hypotheses.}
$H_i$ is calculated on the base of identifying the matched unit $a_i^*$ for unit $i$ in target image. Therefore, if the estimated $a_i^*$ significantly deviates from the ground truth $\overline{a_i^*}$, $H_i$ would be entirely incorrect. In such conditions, we use classification loss to enhance the feature similarity between $a_i^*$ and $\overline{a_i^*}$. In the opposite case, directly supervising the point correspondences calculated through $H_i$ yields better results. The methodology is detailed as follows:

\begin{equation}
\begin{aligned}
  q_1 &=\{i| {\theta}_1 < |a_i^* - \hat{a}_i^*|_{\infty}\}, \\
  q_2 &=\{i| {\theta}_1 \geq |a_i^* - \hat{a}_i^*|_{\infty}\}, \\
    L_H &= 
\begin{cases} 
1 - CosSim(f^1_i,f^1_{a_i^*}) & , i \in q_1, \\
\sum\limits_{p \in P_i}{e_p} & , i \in q_2, \\
\end{cases}
  \label{eq:first_classify}
\end{aligned}
\end{equation} 
where $|*|_{\infty}$  denotes the computation of infinity norm, $CosSim$ is the cosine similarity of two features in the feature map $M_1$. $P_i$ is the 
set of sampled points $p$ that apply $H_i$ as $\overline{H}_p$.

\noindent\textbf{Two-stage Training Process.}
In the entire feature matching process, we divide the training process into two stages, the coarse stage and the fine stage. The fine stage will freeze all the parameters of the coarse stage during training. The coarse stage includes the homography hypotheses estimation, while the fine stage includes segmentation and refinement. The losses used in these two parts are:

\begin{equation}
\begin{aligned}
  &L_{coarse} = L_H,\\
  &L_{fine} = L_s + L_r.
  \label{eq:supervision}
\end{aligned}
\end{equation}

\subsection{Implementation Details}
For feature extracting, we use Resnet-18~\cite{resnet}, then we perform transformer~\cite{transformer} five times at $M_1$. We implement uni-directional cross-attention once in the process of refinement. We 
train our outdoor model and indoor model respectively. The outdoor model is trained on the Megadepth~\cite{megadepth} dataset, while the indoor model is trained on a mixed dataset of Megadepth and ScanNet~\cite{scannet}. The training process is divided into three stages: the first stage is training on data of 640x480 resolution; in the second stage, the longer side is scaled to 640, and some images are rotated by 90 degrees for adaptation training for the coarse; the third stage involves training the fine on 640x480 data. The learning rate used in the first stage is 1e-4. In the second stage is 5e-5, and in the third stage is 3e-4. Both models are trained using three RTX 3090 for 80 hours, with a batch size of 24 in the first stage and 16 in both the second and third stages. We perform all inferences using PyTorch, merely following the implementation of LightGlue to pre-compile the transformer in the coarse stage with PyTorch.

\section{Experiment}

We conducted these evaluations on four different datasets for outdoor and indoor relative pose estimation and homography estimation. These experiments demonstrate superior performance on various downstream tasks.

\subsection{Homography Estimation}
As our first experiment, we evaluate our quality of correspondences and the ability to fit the homography matrix for planar scenes on the HPatches~\cite{hpatches} dataset.

\noindent\textbf{Experimental Setup.}
We conducted comparative experiments using the image matching toolbox proposed by \cite{patch2pix}. Our experiments were configured to replicate the settings outlined for SuperPoint\cite{superpoint}, SuperGlue~\cite{superglue}, and LoFTR~\cite{loftr} as shown in this toolbox. For LightGlue~\cite{lightglue}, we follow their open-source code settings.
To estimate the homography, we employed the RANSAC algorithm with a threshold of 0.25 pixels, leveraging the OpenCV library.
To comprehensively assess the performance of each method, we considered three key metrics: the proportion of matched points with an error within a 1-pixel threshold, the average corner distance for estimated homography matrices measuring less than 1/3/5 pixels, and the average computational time. These metrics were chosen to simultaneously evaluate the matching accuracy, homography estimation precision, and computational efficiency of the methods.
We perform this experiment on a RTX2070 GPU, and we turn off all acceleration options for pytorch implementations, such as flash attention and  precompilation. In order to get as close as possible to a real usage scenario, here, we do not use a warm-up operation when measuring the computing speed.

\noindent\textbf{Dataset.}
HPatches~\cite{hpatches} contains 52 sequences under significant illumination changes and 56 sequences that exhibit large variations in viewpoints. All images are resized with longer dimensions equal to 640.

\noindent\textbf{Results.}
We compare ETO with SuperPoint~\cite{superpoint}, SuperGlue~\cite{superglue}, LightGlue~\cite{lightglue} and LoFTR~\cite{loftr}. 
According to Table.~\ref{tab:hpatches}, our experimental results demonstrate that our method excels in homography estimation accuracy compared to SuperGlue, achieving lower errors within a 1-pixel threshold when compared to both SuperGlue and LightGlue. Furthermore, our approach is significantly faster, outpacing all other methods several times in the evaluation.

\subsection{Outdoor Pose Estimation}
\label{sec:pose_estimation}
We assess the efficacy of our approach for relative pose estimation in the same setting using two distinct datasets: YFCC100M~\cite{yfcc100m} and Megadepth~\cite{megadepth}  for outdoor scenes. 

\begin{table}[!t]
\vspace{-1.5cm}

\tabcolsep 4pt
\footnotesize
% \normalsize

\centering
\begin{tabular}{cccccc}
\toprule
\multirow{2}{*}{Methods} & \multicolumn{3}{c}{Average Corner Error} &Point Accuracy & time     \\ \cline{2-6} 
       & 1px(\%)            & 3px(\%)           & 5px(\%) & 1px(\%)  &  ms        \\ \hline
\multicolumn{1}{l|}{LoFTR~\cite{loftr}}                                           & 46         & 77          & 86  &63    & 218     \\
\multicolumn{1}{l|}{SP~\cite{superpoint}+LG~\cite{lightglue}}                                        & 44          & 73          & 85  &51 & 101        \\
\multicolumn{1}{l|}{SP~\cite{superpoint}+SG~\cite{superglue}}              & 41          & 72          & 82    & 47 & 79      \\ 
\multicolumn{1}{l|}{SP~\cite{superpoint}+search}              & 38          &    68       & 81  &  32 & 81      \\
  \hline
\multicolumn{1}{l|}{Ours}                                                         & 42 & 72 & 82 & 52 &\textbf{53} \\
\bottomrule
\end{tabular}
% \vspace{-0.3cm}
\caption{Evaluation on HPatches~\cite{hpatches} for homography estimation.}
\vspace{-0.4cm}
\label{tab:hpatches}

\end{table}

\begin{table}[!t]
\vspace{-1.0cm}

\tabcolsep 4pt
\footnotesize
% \normalsize

\centering
\begin{tabular}{ccccccccc}
\toprule
\multirow{2}{*}{Methods} & \multicolumn{4}{c}{Megadepth}  & \multicolumn{4}{c}{YFCC100M}    \\ \cline{2-5} \cline{6-9}
       & @5\textdegree            & @10\textdegree           & @20\textdegree  &  ms & @5\textdegree            & @10\textdegree           & @20\textdegree  &  ms        \\ \hline
\multicolumn{1}{l|}{ASpanFormer~\cite{aspanformer}}                                & 58.6    & 72.2    &  81.7   & 158.5   & 44.5    & 63.5    &  78.1   & 155.5  \\
\multicolumn{1}{l|}{Quadtree~\cite{quadtree}}&                                58.6     &    72.1       &    81.5       &   147.6 &     44.7      &   63.9        &      78.2 & 159.4       \\
\multicolumn{1}{l|}{LoFTR~\cite{loftr}}                                           & 57.5          & 71.2          & 80.8     & 93.2  & 44.7          & 63.6          & 78.3     & 96.3    \\
\multicolumn{1}{l|}{SP~\cite{superpoint}+LG~\cite{lightglue}}                                        & 51.5          & 67.7          & 78.9  & 64.2    & 36.1          & 56.2          & 73.1  & 60.8      \\
\multicolumn{1}{l|}{SP~\cite{superpoint}+LG*~\cite{lightglue}}                                        & 47.1          & 64.0          & 77.3  & 26.9   & 29.2          & 48.8          & 67.0  & 27.2      \\
\multicolumn{1}{l|}
{SP~\cite{superpoint}+SG~\cite{superglue}}              & 43.2          & 60.0          & 72.8    & 43.9    & 29.7          & 49.6          & 67.9    & 48.7  \\ 
\multicolumn{1}{l|}{SP~\cite{superpoint}+search}              & 28.8          &    43.4       & 56.6  &  23.7   & 14.0          &    27.0       & 42.2  &  24.6    \\
\multicolumn{1}{l|}{RoMa~\cite{roma}}              & 64.8          &    77.4       & 86.1  &  689   & *          &    *       & *  &  *    \\
\multicolumn{1}{l|}{Tiny-RoMa~\cite{roma}}              & 36.2          &    53.6       & 67.5  &  29.0   & *          &    *       & *  &  *    \\
\hline
\multicolumn{1}{l|}{Ours}                                                         & 51.7 & 66.6 & 77.4 &\textbf{21.0}  & \textbf{44.8} & \textbf{64.0} & \textbf{78.8} &\textbf{22.1} \\
\bottomrule
\end{tabular}
% \vspace{-0.3cm}
\caption{Evaluation on Megadepth~\cite{megadepth} and YFCC100M~\cite{yfcc100m} for outdoor pose estimation.}
\vspace{-0.7cm}
\label{tab:megadepth}

\end{table}

\noindent\textbf{Experimental setup.}
We report the pose accuracy in terms of AUC metric at multiple thresholds ($5^\circ$,$10^\circ$,$20^\circ$) and runtime for every approach, and the RANSAC threshold here is set as 0.25 pixel for all methods. All of the evaluations here are conducted on a RTX2080ti. We turn on flash-attention for LightGlue and turn on the pre-compilation to accelerate the transformer for LightGlue and ETO. Here LightGlue~\cite{lightglue} is slower than SuperGlue~\cite{superglue} for the reason that following the default configuration LightGlue extracts 2048 keypoints and resizes the resolution of images to 1024, while SuperGlue extracts only 1024 keypoints and  keep the resolution of images as the same. 
LightGlue* apply the setting of SuperGlue.
It is imperative to highlight that our method encompasses 4800 points to be matched here, which is the same as LoFTR~\cite{loftr}, ASpanFormer~\cite{aspanformer}, and Quadtree~\cite{quadtree}. To ensure an accurate representation of the actual computation speed, we initiate a warm-up phase for each method, consisting of 10 iterations, prior to conducting measurements.
% More details are provided in the supplementary material.

\noindent\textbf{Dataset.}
YFCC100M~\cite{yfcc100m} encompasses an extensive repository comprising 100 million media assets. For our evaluation, we following~\cite{superglue} and focus on a subset of YFCC100M, specifically four handpicked image collections featuring prominent landmarks, in accordance with the criteria outlined in \cite{superglue} and \cite{loftr}.
MegaDepth, on the other hand, comprises a dataset containing one million Internet-sourced images depicting 196 distinct outdoor scenes. To ensure the integrity of our evaluation protocol, in line with the guidelines presented in~\cite{pats}, we randomly select 1000 image pairs, guaranteeing that none of these pairs have been used in the training processes of any existing methods.
All images in Megadepth and YFCC100M are resized with a resolution equal to 640*480.

\noindent\textbf{Results.}
We compare ETO with SuperPoint~\cite{superpoint}, SuperGlue~\cite{superglue}, LightGlue~\cite{lightglue}, LoFTR~\cite{loftr}, ASpanFormer~\cite{aspanformer} and Quadtree~\cite{quadtree}. 
According to the results shown in Table~\ref{tab:megadepth}, on the easier outdoor cases in MegaDepth, the accuracy of our method for pose estimation is lower than advanced detector-free method but is higher than any detector-based approaches, while our runtime is at most 23\% of the detector-free methods, 81\% of the detector-based methods and 90\% of the CNN-based methods.
While on the more difficult outdoor cases in YFCC100M, the performance of our model is much better than detector-based methods and is comparable with detector-free methods. And still, our superiority on runtime is preserved.
% We show the qualitative result in Fig.~\ref{fig:results}.
% and provide more results in the supplementary materials.

\subsection{Indoor Pose Estimation}
We evaluate our method for indoor pose estimation with ScanNet-1500~\cite{scannet} following~\cite{superglue,loftr}.

\noindent\textbf{Experimental Setup.}
Just like outdoor cases, we report the pose accuracy in terms of the AUC metric at multiple thresholds ($5^\circ$,$10^\circ$,$20^\circ$) and runtime for every approach. However, here we set all of the RANSAC thresholds as 0.5 pixels. All of the images are resized with longer dimensions equal to 640. This evaluation is conducted on RTX2080ti. We have done a warm-up here in measuring the efficiency.
% More details are provided in the supplementary material.

\noindent\textbf{Dataset.}
The ScanNet dataset represents a comprehensive indoor RGB-D collection encompassing 1,613 distinct sequences that cumulatively offer 2.5 million unique views. Each view within this dataset is meticulously annotated with a corresponding ground truth camera pose and depth map. We follow the same training and testing split used by \cite{superglue}.

% \begin{figure*}[t]
% %\vspace{-0.6cm}
% \begin{center}
% \includegraphics[width=1.00\textwidth]{images/results.png}
% \end{center}
% % \vspace{-0.6cm}
%    \caption{\textbf{Qualitative Results of Feature Matching.} Inlier matches are highlighted in green and outliers in red. For visual clarity, the displayed matches are reduced to one-tenth of the actual number. As can be seen from the figure, our method is robust to various extreme scenarios and thus can achieve very superior performance.}
% \vspace{-0.4cm}
% \label{fig:results}
% \end{figure*}

\noindent\textbf{Results.}
We compare our approach with SuperPoint~\cite{superpoint}, SuperGlue~\cite{superglue}, LoFTR~\cite{loftr}, ASpanFormer~\cite{aspanformer} and Quadtree~\cite{quadtree}.
The results are demonstrated in Table~\ref{tab:scannet}. We find that our results are comparable with LoFTR and are superior to SuperPoint+search and SuperPoint+SuperGlue, while much faster than any other methods.
% We show the qualitative result in Fig.~\ref{fig:results}.
% and provide more results in the supplementary materials.

\subsection{Ablation Studies}
\label{sec:ablation}
To evaluate the impact of each design component on the overall structure, we perform an ablation study using the MegaDepth dataset. We systematically add each design element one at a time. The quantitative results are presented in Table~\ref{tab:ablation}.

\noindent\textbf{Base32 w/o Homography} 
We match the units on $M_1$ at H/32 * W/32 resolution and permit the target of unit centroid to be continuous, and we can compute it as Section.~\ref{sec:hypotheses}, while other parameters for the unit are still fixed. We output four virtual correspondences as matches. While it offers rapid processing, it does not achieve a high level of accuracy.

\noindent\textbf{Base8 w/o Homography} 
We set this ablation experiment as the coarse matching of LoFTR~\cite{loftr}. Here we match every possible 8*8 units. We output the center of corresponding units as matches. It is more accurate but too slow.

\noindent\textbf{Base32 w/ Homography.} 
Following Section.~\ref{sec:hypotheses}, we estimate the whole homography matrix and output four virtual correspondences as matches. It performs better than the coarse matching of LoFTR~\cite{loftr} while providing higher efficiency at the same time, 

\noindent\textbf{Basic Refinement w/ Segmentation.}
Following ~\cite{loftr}, we set a layer of transformer between 25 tokens on these two images and try to refine our results while the transformer is trained for 12 hours, which is the same as the training time of our uni-directional attention for the refinement stage. While full attention execution speed is considerably slower than that of uni-directional attention, its accuracy is merely comparable with the latter. 

\noindent\textbf{Uni-directional w/o Segmentation.}
Here we directly choose the homography hypotheses $H_5$ for each unit $j$ which is in the center. Then, we conduct the refinement as the same. The results show that the segmentation stage significantly improves the accuracy.

\begin{table}[!t]
\vspace{-1.0cm}

\tabcolsep 4pt
\footnotesize
% \normalsize

\centering
\begin{tabular}{ccccc}
\toprule
\multirow{2}{*}{Methods} & \multicolumn{3}{c}{Pose estimation AUC} & average time     \\ \cline{2-5} 
       & @5\textdegree            & @10\textdegree           & @20\textdegree  &  ms        \\ \hline
\multicolumn{1}{l|}{ASpanFormer~\cite{aspanformer}}                                & 24.5    & 45.0    &  62.8   & 160.0  \\ 
\multicolumn{1}{l|}{Quadtree~\cite{quadtree}}                                     &    23.9       &     43.0      &     60.2 & 145.9      \\
\multicolumn{1}{l|}{LoFTR~\cite{loftr}}                                           & 21.4          & 40.3          & 57.2     & 94.2     \\
\multicolumn{1}{l|}{SuperPoint~\cite{superpoint}+SuperGlue~\cite{superglue}}              & 13.7          & 29.8          & 47.2    & 63.1      \\ 
\multicolumn{1}{l|}{Superpoint~\cite{superpoint}+search}              & 8.0          &    18.3       & 29.8  &  27.1      \\
\hline
\multicolumn{1}{l|}{Ours}                                                         & 20.1 & 40.4 & 59.8 &\textbf{24.2} \\
\bottomrule
\end{tabular}
% \vspace{-0.3cm}
\caption{Evaluation on Scannet~\cite{scannet} for indoor pose estimation.}
\vspace{-0.3cm}
\label{tab:scannet}

\end{table}

\begin{table}[!t]
\vspace{-0.3cm}

\tabcolsep 4pt
\footnotesize
% \normalsize

\centering
\begin{tabular}{ccccc}
\toprule
\multirow{2}{*}{Ablation} & \multicolumn{3}{c}{Pose estimation AUC} & time     \\ \cline{2-5} 
       & @5\textdegree            & @10\textdegree           & @20\textdegree  &  ms        \\ \hline
\multicolumn{1}{l|}{Base32 w/o Homography}              & 9.5          &    21.3       & 36.3  &  8.5      \\
\multicolumn{1}{l|}{Base8 w/o Homography~(LoFTR coarse)}   & 25.7      & 41.8    & 57.7   & 58.9      \\ 
\multicolumn{1}{l|}{Base32 w/ Homography}                                        & 28.5          & 46.2          & 61.4  & 8.5        \\
\multicolumn{1}{l|}{Basic Refinement w/ segmentation}                                     &   51.0        &    66.6       &    77.6 &   32.8    \\
\multicolumn{1}{l|}{Uni-directional w/o segmentation}                                           & 42.1          & 59.0          & 72.0     & 21.2    \\
\multicolumn{1}{l|}{Full}                                & 51.7    & 66.6    &  77.4   & 22.0  \\ 
\bottomrule
\end{tabular}
% \vspace{-0.3cm}
\caption{Ablation study based on Megadepth~\cite{megadepth} for outdoor pose estimation.}
\vspace{-0.5cm}
\label{tab:ablation}

\end{table}

\section{Conclusion and Limitations.}
\label{sec:conclusion}
In this paper, we propose Efficient Transformer-based Local Feature Matching by Organizing Multiple Homography hypotheses (ETO). ETO tries to approximate a continuous corresponding function with multiple homography hypotheses with fewer tokens fed to the transformer. Multiple datasets demonstrate that ETO delivers nearly comparable performance in relative pose estimation and homography estimation with other transformer-based methods, while its speed surpasses all of them by a large margin. However, there remains significant space for improvement in ETO's matching accuracy. Next, we could explore an end-to-end training mode, which would allow for further enhancement of the feature extractor at a fine-grained level. Moreover, we believe that intermediate-level features can provide not only segmentation information but also data conducive to more precise matching. Finally, the form of parametric scheme we present here may not be optimal and complete for homography transformation, so we will continue to explore better parametric schemes. These strategies are expected to enable our method to compete with approaches like PATS~\cite{pats} and DKM~\cite{dkm} in terms of matching precision, without considerably compromising speed. 
\noindent\textbf{Acknowledgements.}
This work was partially supported by NSF of China  (No. 61932003).

\clearpage

\bibliographystyle{unsrtnat}
\bibliography{main}

\begin{thebibliography}{62}
\providecommand{\natexlab}[1]{#1}
\providecommand{\url}[1]{\texttt{#1}}
\expandafter\ifx\csname urlstyle\endcsname\relax
  \providecommand{\doi}[1]{doi: #1}\else
  \providecommand{\doi}{doi: \begingroup \urlstyle{rm}\Url}\fi

\bibitem[Rublee et~al.(2011)Rublee, Rabaud, Konolige, and Bradski]{orb}
Ethan Rublee, Vincent Rabaud, Kurt Konolige, and Gary~R. Bradski.
\newblock {ORB:} an efficient alternative to {SIFT} or {SURF}.
\newblock In \emph{{Proceedings of the IEEE/CVF International Conference on Computer Vision}}, pages 2564--2571. IEEE, 2011.

\bibitem[Bay et~al.(2008)Bay, Ess, Tuytelaars, and Gool]{surf}
Herbert Bay, Andreas Ess, Tinne Tuytelaars, and Luc~Van Gool.
\newblock Speeded-up robust features {(SURF)}.
\newblock \emph{{Comput. Vis. Image Underst.}}, 110\penalty0 (3):\penalty0 346--359, 2008.

\bibitem[Mur{-}Artal et~al.(2015)Mur{-}Artal, Montiel, and Tard{\'{o}}s]{orb-slam}
Raul Mur{-}Artal, J.~M.~M. Montiel, and Juan~D. Tard{\'{o}}s.
\newblock {ORB-SLAM:} {A} versatile and accurate monocular {SLAM} system.
\newblock \emph{{IEEE} Trans. Robotics}, 31\penalty0 (5):\penalty0 1147--1163, 2015.

\bibitem[Yang et~al.(2022{\natexlab{a}})Yang, Li, Zhai, Ming, Liu, and Zhang]{yang2022vox}
Xingrui Yang, Hai Li, Hongjia Zhai, Yuhang Ming, Yuqian Liu, and Guofeng Zhang.
\newblock Vox-fusion: Dense tracking and mapping with voxel-based neural implicit representation.
\newblock In \emph{{IEEE International Symposium on Mixed and Augmented Reality}}, pages 499--507. IEEE, 2022{\natexlab{a}}.

\bibitem[Liu et~al.(2023)Liu, Li, Teng, Bao, Zhang, Zhang, and Cui]{tof_slam}
Xinyang Liu, Yijin Li, Yanbin Teng, Hujun Bao, Guofeng Zhang, Yinda Zhang, and Zhaopeng Cui.
\newblock Multi-modal neural radiance field for monocular dense slam with a light-weight tof sensor.
\newblock In \emph{Proceedings of the ieee/cvf international conference on computer vision}, pages 1--11, 2023.

\bibitem[Hu et~al.(2024{\natexlab{a}})Hu, Chen, Feng, Li, Yang, Bao, Zhang, and Cui]{hu2024cg}
Jiarui Hu, Xianhao Chen, Boyin Feng, Guanglin Li, Liangjing Yang, Hujun Bao, Guofeng Zhang, and Zhaopeng Cui.
\newblock Cg-slam: Efficient dense rgb-d slam in a consistent uncertainty-aware 3d gaussian field.
\newblock \emph{arXiv preprint arXiv:2403.16095}, 2024{\natexlab{a}}.

\bibitem[Hu et~al.(2024{\natexlab{b}})Hu, Mao, Bao, Zhang, and Cui]{hu2024cp}
Jiarui Hu, Mao Mao, Hujun Bao, Guofeng Zhang, and Zhaopeng Cui.
\newblock Cp-slam: Collaborative neural point-based slam system.
\newblock \emph{Advances in Neural Information Processing Systems}, 36, 2024{\natexlab{b}}.

\bibitem[Zhai et~al.(2024{\natexlab{a}})Zhai, Huang, Hu, Li, Bao, and Zhang]{nis_slam}
Hongjia Zhai, Gan Huang, Qirui Hu, Guanglin Li, Hujun Bao, and Guofeng Zhang.
\newblock Nis-slam: Neural implicit semantic rgb-d slam for 3d consistent scene understanding.
\newblock \emph{IEEE Transactions on Visualization and Computer Graphics}, pages 1--11, 2024{\natexlab{a}}.

\bibitem[Chen et~al.(2024)Chen, Peng, Li, Ju, Bao, Chen, and Zhang]{mvn_afm}
Shuo Chen, Mao Peng, Yijin Li, Bing-Feng Ju, Hujun Bao, Yuan-Liu Chen, and Guofeng Zhang.
\newblock Multi-view neural 3d reconstruction of micro-and nanostructures with atomic force microscopy.
\newblock \emph{Communications Engineering}, 3\penalty0 (1):\penalty0 131, 2024.

\bibitem[Yang et~al.(2022{\natexlab{b}})Yang, Zhang, Li, Cui, Fanello, Bao, and Zhang]{nr_in_a_room}
Bangbang Yang, Yinda Zhang, Yijin Li, Zhaopeng Cui, Sean Fanello, Hujun Bao, and Guofeng Zhang.
\newblock Neural rendering in a room: amodal 3d understanding and free-viewpoint rendering for the closed scene composed of pre-captured objects.
\newblock \emph{ACM Transactions on Graphics (TOG)}, 41\penalty0 (4):\penalty0 1--10, 2022{\natexlab{b}}.

\bibitem[Sarlin et~al.(2019{\natexlab{a}})Sarlin, Cadena, Siegwart, and Dymczyk]{hloc}
Paul{-}Edouard Sarlin, Cesar Cadena, Roland Siegwart, and Marcin Dymczyk.
\newblock From coarse to fine: Robust hierarchical localization at large scale.
\newblock In \emph{{Proceedings of the IEEE/CVF Conference on Computer Vision and Pattern Recognition}}, pages 12716--12725, 2019{\natexlab{a}}.

\bibitem[Huang et~al.(2021)Huang, Zhou, Li, Yang, Xu, Zhou, Bao, Zhang, and Li]{vs_net}
Zhaoyang Huang, Han Zhou, Yijin Li, Bangbang Yang, Yan Xu, Xiaowei Zhou, Hujun Bao, Guofeng Zhang, and Hongsheng Li.
\newblock Vs-net: Voting with segmentation for visual localization.
\newblock In \emph{Proceedings of the IEEE/CVF Conference on Computer Vision and Pattern Recognition}, pages 6101--6111, 2021.

\bibitem[Zhai et~al.(2024{\natexlab{b}})Zhai, Zhang, Boming, Li, He, Cui, Bao, and Zhang]{splatloc}
Hongjia Zhai, Xiyu Zhang, Zhao Boming, Hai Li, Yijia He, Zhaopeng Cui, Hujun Bao, and Guofeng Zhang.
\newblock Splatloc: 3d gaussian splatting-based visual localization for augmented reality.
\newblock \emph{arXiv preprint arXiv:2409.14067}, 2024{\natexlab{b}}.

\bibitem[Xu et~al.(2022)Xu, Lin, Zhang, Wang, and Li]{rnnpose}
Yan Xu, Kwan{-}Yee Lin, Guofeng Zhang, Xiaogang Wang, and Hongsheng Li.
\newblock Rnnpose: Recurrent 6-dof object pose refinement with robust correspondence field estimation and pose optimization.
\newblock In \emph{{Proceedings of the IEEE/CVF Conference on Computer Vision and Pattern Recognition}}, pages 14880--14890, 2022.

\bibitem[Li et~al.(2022{\natexlab{a}})Li, Li, Ye, Zhang, Kong, Cui, and Zhang]{gcasp}
Guanglin Li, Yifeng Li, Zhichao Ye, Qihang Zhang, Tao Kong, Zhaopeng Cui, and Guofeng Zhang.
\newblock Generative category-level shape and pose estimation with semantic primitives.
\newblock In \emph{Conference on Robot Learning}, pages 1390--1400. PMLR, 2022{\natexlab{a}}.

\bibitem[DeTone et~al.(2018)DeTone, Malisiewicz, and Rabinovich]{superpoint}
Daniel DeTone, Tomasz Malisiewicz, and Andrew Rabinovich.
\newblock Superpoint: Self-supervised interest point detection and description.
\newblock In \emph{{Proceedings of the {IEEE/CVF} Conference on Computer Vision and Pattern Recognition}}, pages 224--236, 2018.

\bibitem[Dusmanu et~al.(2019)Dusmanu, Rocco, Pajdla, Pollefeys, Sivic, Torii, and Sattler]{d2net}
Mihai Dusmanu, Ignacio Rocco, Tom{\'{a}}s Pajdla, Marc Pollefeys, Josef Sivic, Akihiko Torii, and Torsten Sattler.
\newblock D2-net: {A} trainable {CNN} for joint description and detection of local features.
\newblock In \emph{{Proceedings of the IEEE/CVF Conference on Computer Vision and Pattern Recognition}}, pages 8092--8101, 2019.

\bibitem[Ni et~al.(2023)Ni, Li, Huang, Li, Bao, Cui, and Zhang]{pats}
Junjie Ni, Yijin Li, Zhaoyang Huang, Hongsheng Li, Hujun Bao, Zhaopeng Cui, and Guofeng Zhang.
\newblock Pats: Patch area transportation with subdivision for local feature matching.
\newblock In \emph{Proceedings of the IEEE/CVF Conference on Computer Vision and Pattern Recognition}, pages 17776--17786, 2023.

\bibitem[Chen et~al.(2022)Chen, Luo, Zhou, Tian, Zhen, Fang, Mckinnon, Tsin, and Quan]{aspanformer}
Hongkai Chen, Zixin Luo, Lei Zhou, Yurun Tian, Mingmin Zhen, Tian Fang, David Mckinnon, Yanghai Tsin, and Long Quan.
\newblock Aspanformer: Detector-free image matching with adaptive span transformer.
\newblock In \emph{{European Conference on Computer Vision}}, 2022.

\bibitem[Sun et~al.(2021)Sun, Shen, Wang, Bao, and Zhou]{loftr}
Jiaming Sun, Zehong Shen, Yuang Wang, Hujun Bao, and Xiaowei Zhou.
\newblock Loftr: Detector-free local feature matching with transformers.
\newblock In \emph{{Proceedings of the IEEE/CVF Conference on Computer Vision and Pattern Recognition}}, pages 8922--8931, 2021.

\bibitem[Vaswani et~al.(2017)Vaswani, Shazeer, Parmar, Uszkoreit, Jones, Gomez, Kaiser, and Polosukhin]{transformer}
Ashish Vaswani, Noam Shazeer, Niki Parmar, Jakob Uszkoreit, Llion Jones, Aidan~N. Gomez, Lukasz Kaiser, and Illia Polosukhin.
\newblock Attention is all you need.
\newblock \emph{{Advances in Neural Information Processing Systems}}, 30, 2017.

\bibitem[Katharopoulos et~al.(2020)Katharopoulos, Vyas, Pappas, and Fleuret]{linear_transformer}
Angelos Katharopoulos, Apoorv Vyas, Nikolaos Pappas, and Fran{\c{c}}ois Fleuret.
\newblock Transformers are rnns: Fast autoregressive transformers with linear attention.
\newblock In \emph{International conference on machine learning}, pages 5156--5165. PMLR, 2020.

\bibitem[Harley et~al.(2022)Harley, Fang, and Fragkiadaki]{tracking}
Adam~W Harley, Zhaoyuan Fang, and Katerina Fragkiadaki.
\newblock Particle video revisited: Tracking through occlusions using point trajectories.
\newblock In \emph{European Conference on Computer Vision}, pages 59--75. Springer, 2022.

\bibitem[Li et~al.(2024)Li, Shen, Huang, Chen, Bian, Shi, Wang, Sun, Bao, Cui, Zhang, and Li]{blinkvision}
Yijin Li, Yichen Shen, Zhaoyang Huang, Shuo Chen, Weikang Bian, Xiaoyu Shi, Fu-Yun Wang, Keqiang Sun, Hujun Bao, Zhaopeng Cui, Guofeng Zhang, and Hongsheng Li.
\newblock Blinkvision: A benchmark for optical flow, scene flow and point tracking estimation using rgb frames and events.
\newblock In \emph{European conference on computer vision}. Springer, 2024.

\bibitem[Bian et~al.(2024)Bian, Huang, Shi, Dong, Li, and Li]{context_tap}
Weikang Bian, Zhaoyang Huang, Xiaoyu Shi, Yitong Dong, Yijin Li, and Hongsheng Li.
\newblock Context-pips: Persistent independent particles demands context features.
\newblock \emph{Advances in Neural Information Processing Systems}, 36, 2024.

\bibitem[Sarlin et~al.(2019{\natexlab{b}})Sarlin, Cadena, Siegwart, and Dymczyk]{large_scale}
Paul-Edouard Sarlin, Cesar Cadena, Roland Siegwart, and Marcin Dymczyk.
\newblock From coarse to fine: Robust hierarchical localization at large scale.
\newblock In \emph{Proceedings of the IEEE/CVF Conference on Computer Vision and Pattern Recognition}, pages 12716--12725, 2019{\natexlab{b}}.

\bibitem[Barnes et~al.(2009)Barnes, Shechtman, Finkelstein, and Goldman]{patchmatch}
Connelly Barnes, Eli Shechtman, Adam Finkelstein, and Dan~B Goldman.
\newblock Patchmatch: A randomized correspondence algorithm for structural image editing.
\newblock \emph{ACM Trans. Graph.}, 28\penalty0 (3):\penalty0 24, 2009.

\bibitem[Redmon and Farhadi(2018)]{yolov3}
Joseph Redmon and Ali Farhadi.
\newblock Yolov3: An incremental improvement.
\newblock \emph{arXiv preprint arXiv:1804.02767}, 2018.

\bibitem[Lindenberger et~al.(2023)Lindenberger, Sarlin, and Pollefeys]{lightglue}
Philipp Lindenberger, Paul-Edouard Sarlin, and Marc Pollefeys.
\newblock {LightGlue: Local Feature Matching at Light Speed}.
\newblock In \emph{ICCV}, 2023.

\bibitem[Li and Snavely(2018)]{megadepth}
Zhengqi Li and Noah Snavely.
\newblock Megadepth: Learning single-view depth prediction from internet photos.
\newblock In \emph{{Proceedings of the IEEE/CVF Conference on Computer Vision and Pattern Recognition}}, 2018.

\bibitem[Thomee et~al.(2016)Thomee, Shamma, Friedland, Elizalde, Ni, Poland, Borth, and Li]{yfcc100m}
Bart Thomee, David~A. Shamma, Gerald Friedland, Benjamin Elizalde, Karl Ni, Douglas Poland, Damian Borth, and Li{-}Jia Li.
\newblock {YFCC100M:} the new data in multimedia research.
\newblock \emph{Commun. {ACM}}, 59\penalty0 (2):\penalty0 64--73, 2016.

\bibitem[Dai et~al.(2017)Dai, Chang, Savva, Halber, Funkhouser, and Nie{\ss}ner]{scannet}
Angela Dai, Angel~X. Chang, Manolis Savva, Maciej Halber, Thomas~A. Funkhouser, and Matthias Nie{\ss}ner.
\newblock Scannet: Richly-annotated 3d reconstructions of indoor scenes.
\newblock In \emph{{Proceedings of the IEEE/CVF Conference on Computer Vision and Pattern Recognition}}, pages 5828--5839, 2017.

\bibitem[Balntas et~al.(2017)Balntas, Lenc, Vedaldi, and Mikolajczyk]{hpatches}
Vassileios Balntas, Karel Lenc, Andrea Vedaldi, and Krystian Mikolajczyk.
\newblock Hpatches: A benchmark and evaluation of handcrafted and learned local descriptors.
\newblock In \emph{CVPR}, 2017.

\bibitem[Lowe(2004)]{sift}
David~G. Lowe.
\newblock Distinctive image features from scale-invariant keypoints.
\newblock \emph{Int. J. Comput. Vis.}, 60\penalty0 (2):\penalty0 91--110, 2004.

\bibitem[Revaud et~al.(2019)Revaud, de~Souza, Humenberger, and Weinzaepfel]{r2d2}
J{\'{e}}r{\^{o}}me Revaud, C{\'{e}}sar~Roberto de~Souza, Martin Humenberger, and Philippe Weinzaepfel.
\newblock {R2D2:} reliable and repeatable detector and descriptor.
\newblock \emph{{Advances in Neural Information Processing Systems}}, 32, 2019.

\bibitem[Sarlin et~al.(2020)Sarlin, DeTone, Malisiewicz, and Rabinovich]{superglue}
Paul{-}Edouard Sarlin, Daniel DeTone, Tomasz Malisiewicz, and Andrew Rabinovich.
\newblock Superglue: Learning feature matching with graph neural networks.
\newblock In \emph{{Proceedings of the IEEE/CVF Conference on Computer Vision and Pattern Recognition}}, pages 4938--4947, 2020.

\bibitem[Yi et~al.(2018)Yi, Trulls, Ono, Lepetit, Salzmann, and Fua]{goodcorr}
Kwang~Moo Yi, Eduard Trulls, Yuki Ono, Vincent Lepetit, Mathieu Salzmann, and Pascal Fua.
\newblock Learning to find good correspondences.
\newblock In \emph{{Proceedings of the IEEE/CVF Conference on Computer Vision and Pattern Recognition}}, pages 2666--2674, 2018.

\bibitem[Jiang et~al.(2021)Jiang, Trulls, Hosang, Tagliasacchi, and Yi]{cotr}
Wei Jiang, Eduard Trulls, Jan Hosang, Andrea Tagliasacchi, and Kwang~Moo Yi.
\newblock {COTR:} correspondence transformer for matching across images.
\newblock In \emph{{Proceedings of the IEEE/CVF International Conference on Computer Vision}}, pages 171--180. Springer, 2021.

\bibitem[Tan et~al.(2022)Tan, Liu, Chen, Chen, Zhang, Shen, Ding, and Ji]{ecotr}
Dongli Tan, Jiang-Jiang Liu, Xingyu Chen, Chao Chen, Ruixin Zhang, Yunhang Shen, Shouhong Ding, and Rongrong Ji.
\newblock Eco-tr: Efficient correspondences finding via coarse-to-fine refinement.
\newblock In \emph{European Conference on Computer Vision}, pages 317--334. Springer, 2022.

\bibitem[Li et~al.(2022{\natexlab{b}})Li, Liu, Dong, Zhou, Bao, Zhang, Zhang, and Cui]{deltar}
Yijin Li, Xinyang Liu, Wenqi Dong, Han Zhou, Hujun Bao, Guofeng Zhang, Yinda Zhang, and Zhaopeng Cui.
\newblock {DELTAR:} depth estimation from a light-weight tof sensor and {RGB} image.
\newblock In \emph{{European Conference on Computer Vision}}, pages 619--636. Springer, 2022{\natexlab{b}}.

\bibitem[Huang et~al.(2023)Huang, Shi, Zhang, Wang, Li, Qin, Dai, Wang, and Li]{flowformer}
Zhaoyang Huang, Xiaoyu Shi, Chao Zhang, Qiang Wang, Yijin Li, Hongwei Qin, Jifeng Dai, Xiaogang Wang, and Hongsheng Li.
\newblock Flowformer: A transformer architecture and its masked cost volume autoencoding for optical flow.
\newblock \emph{arXiv preprint arXiv:2306.05442}, 2023.

\bibitem[Li et~al.(2023)Li, Huang, Chen, Shi, Li, Bao, Cui, and Zhang]{blinkflow}
Yijin Li, Zhaoyang Huang, Shuo Chen, Xiaoyu Shi, Hongsheng Li, Hujun Bao, Zhaopeng Cui, and Guofeng Zhang.
\newblock Blinkflow: A dataset to push the limits of event-based optical flow estimation.
\newblock In \emph{2023 IEEE/RSJ International Conference on Intelligent Robots and Systems (IROS)}, pages 3881--3888. IEEE, 2023.

\bibitem[Tang et~al.(2021)Tang, Zhang, Zhu, and Tan]{quadtree}
Shitao Tang, Jiahui Zhang, Siyu Zhu, and Ping Tan.
\newblock Quadtree attention for vision transformers.
\newblock In \emph{The International Conference on Learning Representations}. OpenReview.net, 2021.

\bibitem[Truong et~al.(2021)Truong, Danelljan, Timofte, and Gool]{pdcnet++}
Prune Truong, Martin Danelljan, Radu Timofte, and Luc~Van Gool.
\newblock Pdc-net+: Enhanced probabilistic dense correspondence network.
\newblock \emph{{arXiv preprint arXiv:2109.13912}}, 2021.

\bibitem[Edstedt et~al.(2023)Edstedt, Athanasiadis, Wadenb{\"a}ck, and Felsberg]{dkm}
Johan Edstedt, Ioannis Athanasiadis, M{\aa}rten Wadenb{\"a}ck, and Michael Felsberg.
\newblock Dkm: Dense kernelized feature matching for geometry estimation.
\newblock In \emph{Proceedings of the IEEE/CVF Conference on Computer Vision and Pattern Recognition}, pages 17765--17775, 2023.

\bibitem[Zhang et~al.(2023)Zhang, Sun, Chen, Li, and Shen]{rgm}
Songyan Zhang, Xinyu Sun, Hao Chen, Bo~Li, and Chunhua Shen.
\newblock Rgm: A robust generalist matching model.
\newblock \emph{arXiv preprint arXiv:2310.11755}, 2023.

\bibitem[Teed and Deng(2020)]{raft}
Zachary Teed and Jia Deng.
\newblock {RAFT:} recurrent all-pairs field transforms for optical flow.
\newblock In \emph{{European Conference on Computer Vision}}, pages 402--419. Springer, 2020.

\bibitem[Luo et~al.(2020)Luo, Zhou, Bai, Chen, Zhang, Yao, Li, Fang, and Quan]{aslfeat}
Zixin Luo, Lei Zhou, Xuyang Bai, Hongkai Chen, Jiahui Zhang, Yao Yao, Shiwei Li, Tian Fang, and Long Quan.
\newblock Aslfeat: Learning local features of accurate shape and localization.
\newblock In \emph{{Proceedings of the IEEE/CVF Conference on Computer Vision and Pattern Recognition}}, pages 6589--6598, 2020.

\bibitem[Lindeberg(1998)]{scalespace}
Tony Lindeberg.
\newblock Feature detection with automatic scale selection.
\newblock \emph{{Int. J. Comput. Vis.}}, 30\penalty0 (2):\penalty0 79--116, 1998.

\bibitem[Barroso-Laguna et~al.(2022)Barroso-Laguna, Tian, and Mikolajczyk]{scalenet}
Axel Barroso-Laguna, Yurun Tian, and Krystian Mikolajczyk.
\newblock Scalenet: {A} shallow architecture for scale estimation.
\newblock In \emph{{Proceedings of the IEEE/CVF Conference on Computer Vision and Pattern Recognition}}, pages 12808--12818, 2022.

\bibitem[Fan et~al.(2023)Fan, Chen, Hu, and Zhou]{occ2}
Miao Fan, Mingrui Chen, Chen Hu, and Shuchang Zhou.
\newblock Occ\^{} 2net: Robust image matching based on 3d occupancy estimation for occluded regions.
\newblock In \emph{Proceedings of the IEEE/CVF International Conference on Computer Vision}, pages 9652--9662, 2023.

\bibitem[Dehghani et~al.(2018)Dehghani, Gouws, Vinyals, Uszkoreit, and Kaiser]{universal}
Mostafa Dehghani, Stephan Gouws, Oriol Vinyals, Jakob Uszkoreit, and {\L}ukasz Kaiser.
\newblock Universal transformers.
\newblock \emph{arXiv preprint arXiv:1807.03819}, 2018.

\bibitem[Elbayad et~al.(2019)Elbayad, Gu, Grave, and Auli]{adaptive_transformer}
Maha Elbayad, Jiatao Gu, Edouard Grave, and Michael Auli.
\newblock Depth-adaptive transformer.
\newblock \emph{arXiv preprint arXiv:1910.10073}, 2019.

\bibitem[Wang et~al.(2024)Wang, He, Peng, Tan, and Zhou]{eloftr}
Yifan Wang, Xingyi He, Sida Peng, Dongli Tan, and Xiaowei Zhou.
\newblock Efficient loftr: Semi-dense local feature matching with sparse-like speed.
\newblock \emph{arXiv preprint arXiv:2403.04765}, 2024.

\bibitem[Ronneberger et~al.(2015)Ronneberger, Fischer, and Brox]{unet}
Olaf Ronneberger, Philipp Fischer, and Thomas Brox.
\newblock U-net: Convolutional networks for biomedical image segmentation.
\newblock In \emph{Medical Image Computing and Computer-Assisted Intervention--MICCAI 2015: 18th International Conference, Munich, Germany, October 5-9, 2015, Proceedings, Part III 18}, pages 234--241. Springer, 2015.

\bibitem[He et~al.(2016)He, Zhang, Ren, and Sun]{resnet}
Kaiming He, Xiangyu Zhang, Shaoqing Ren, and Jian Sun.
\newblock Deep residual learning for image recognition.
\newblock In \emph{Proceedings of the IEEE conference on computer vision and pattern recognition}, pages 770--778, 2016.

\bibitem[Cuturi(2013)]{sinkhorn}
Marco Cuturi.
\newblock Sinkhorn distances: Lightspeed computation of optimal transport.
\newblock \emph{{Advances in Neural Information Processing Systems}}, 26, 2013.

\bibitem[Guo et~al.(2019)Guo, Yang, Yang, Wang, and Li]{group_correlation}
Xiaoyang Guo, Kai Yang, Wukui Yang, Xiaogang Wang, and Hongsheng Li.
\newblock Group-wise correlation stereo network.
\newblock In \emph{Proceedings of the IEEE/CVF conference on computer vision and pattern recognition}, pages 3273--3282, 2019.

\bibitem[Lin et~al.(2017)Lin, Goyal, Girshick, He, and Doll{\'a}r]{focal_loss}
Tsung-Yi Lin, Priya Goyal, Ross Girshick, Kaiming He, and Piotr Doll{\'a}r.
\newblock Focal loss for dense object detection.
\newblock In \emph{Proceedings of the IEEE international conference on computer vision}, pages 2980--2988, 2017.

\bibitem[Good(1952)]{bce}
Irving~John Good.
\newblock Rational decisions.
\newblock \emph{Journal of the Royal Statistical Society: Series B (Methodological)}, 14\penalty0 (1):\penalty0 107--114, 1952.

\bibitem[Zhou et~al.(2021)Zhou, Sattler, and Leal{-}Taix{\'{e}}]{patch2pix}
Qunjie Zhou, Torsten Sattler, and Laura Leal{-}Taix{\'{e}}.
\newblock Patch2pix: Epipolar-guided pixel-level correspondences.
\newblock In \emph{{Proceedings of the IEEE/CVF Conference on Computer Vision and Pattern Recognition}}, pages 4669--4678, 2021.

\bibitem[Edstedt et~al.(2024)Edstedt, Sun, B{\"o}kman, Wadenb{\"a}ck, and Felsberg]{roma}
Johan Edstedt, Qiyu Sun, Georg B{\"o}kman, M{\aa}rten Wadenb{\"a}ck, and Michael Felsberg.
\newblock Roma: Robust dense feature matching.
\newblock In \emph{Proceedings of the IEEE/CVF Conference on Computer Vision and Pattern Recognition}, pages 19790--19800, 2024.

\end{thebibliography}

\clearpage

\end{document}